\documentclass[11pt,a4paper]{article}
\usepackage[hyperref]{emnlp2020}
\usepackage{times}
\usepackage{latexsym}
\usepackage{hyperref}
\usepackage{graphicx}
\usepackage{amsmath}
\usepackage{booktabs}
\usepackage{bm}
\usepackage{multicol}
\usepackage{svg}
\usepackage{booktabs}
\usepackage{array}
\usepackage{colortbl}
\usepackage{subcaption}
\usepackage{tikz}
\usepackage{rotating}
\usepackage{enumitem}
\usepackage{pgfplots}
\usepackage{pgfplotstable}

\pgfplotsset{compat=1.16}

\usepackage{microtype}

\aclfinalcopy 

\title{Unbabel's Participation in the WMT20 Metrics Shared Task}

\author{Ricardo Rei \qquad Craig Stewart \qquad Ana C Farinha \qquad Alon Lavie \\ Unbabel AI\\
  {\fontsize{10}{10}\selectfont {\texttt{\{ricardo.rei, craig.stewart, catarina.farinha, alon.lavie\}@unbabel.com}}}  
}

\date{\today}

\begin{document}
\maketitle
\begin{abstract}
We present the contribution of the Unbabel team to the WMT 2020 Shared Task on Metrics. We intend to participate on the segment-level, document-level and system-level tracks on all language pairs, as well as the ``QE as a Metric'' track. Accordingly, we illustrate results of our models in these tracks with reference to test sets from the previous year. Our submissions build upon the recently proposed {\sc Comet} framework: we train several estimator models to regress on different human-generated quality scores and a novel ranking model trained on relative ranks obtained from Direct Assessments. We also propose a simple technique for converting segment-level predictions into a document-level score. Overall, our systems achieve strong results for all language pairs on previous test sets and in many cases set a new state-of-the-art.
\end{abstract}

\section{Introduction}

In this paper we describe our submission to the WMT20 Metrics shared task. Our work is based on the {\sc Comet}\footnote{\textbf{C}rosslingual \textbf{
O}ptimized \textbf{M}etric for \textbf{E}valuation of \textbf{T}ranslation hosted at: \url{https://github.com/Unbabel/COMET}} framework, as presented in \citet{comet}, and extended here to evaluation of MT output at segment, document and system-level, forming the basis of our submissions to the corresponding task tracks. Recently, automatic evaluation of MT has followed most other sub-fields in NLP with a notable interest in leveraging the power of large, pre-trained language models. Metrics such as {\sc Bert Regressor} \citep{shimanaka2019machine}, {\sc Bertscore} \citep{ZhangBERTScore}, {\sc Bleurt}  \citep{sellam-etal-2020-bleurt} and our more recent {\sc Comet} \citep{comet}, all build upon developments in language modelling to generate automatic metrics with high correlation with human judgement. Our MT evaluation models follow a similar strategy, specifically utilizing the most recent iterations of the XLM-RoBERTa model presented in \citet{conneau2019unsupervised}. 

The uniqueness of our approach comes from our inclusion of the source text as input which was demonstrated in \citet{takahashi-etal-2020-automatic} and \citet{comet} to be beneficial to the model. In our contribution to the shared task, we demonstrate methods of further exploiting information in the source text as well as a technique to fully harness the power of pre-trained language models to further improve the prediction accuracy of our evaluation framework when more than one reference translation is available.

For the shared task, we utilize two primary types of models built using the {\sc Comet} framework, namely; the Estimator models, which regress directly on human scores of MT quality such as Direct Assessment; and the {\sc Comet-rank} (base) model used to rank MT outputs and systems.

In addition to the models themselves, we also make the following research contributions:

\begin{enumerate}
    \item We introduce a method for handling multiple references at inference time and for optimizing the utility of information from all available text inputs
    \item We propose a  simple technique for calculating a document-level score from a weighted average of segment-level scores
\end{enumerate}

We demonstrate that our {\sc Comet} framework trained models achieve state-of-the-art results or are competitive on all settings introduced in the WMT19 Metrics shared task, outperforming, in some cases, more recently proposed metrics such as {\sc Bertscore} \citep{ZhangBERTScore}, {\sc Bleurt} \citep{sellam-etal-2020-bleurt} and {\sc Prism} \citep{thompson-post-2020-automatic}.

\section{The {\sc Comet} Framework}

As outlined in \citet{comet}, the {\sc Comet} framework allows for training of specialized evaluation metrics that correlate well with different types of human-generated quality scores. The general structure of the framework consists of a cross-lingual encoder that produces a series of token-level vector embeddings for source, hypothesis and reference inputs, a pooling layer which converts the various token-level representations into segment-level vectors for each input, and a predictive neural network that generates a quality score. The latter model can either be trained to regress directly on a score to produce predictions of segment-level quality, or can be trained as a ranker to differentiate MT systems. In our contribution to the shared task, we introduce two varieties of models built on the {\sc Comet} framework that are extensions of the models evaluated in \citet{comet}.

\section{{\sc Comet} Models}

\subsection{Estimator Models}
Our Estimators generally follow the architecture proposed in \citet{comet}, that is to say we encode segment-level representations using XLM-RoBERTa and pass these outputs through a feed-forward regressor. As in \citet{comet}, we train three versions of this basic estimator model against different types of human judgement; \textit{Human-mediated Translation Edit Rate} (HTER) \cite{Snover06astudy}, a proprietary implementation of \textit{Multidimensional Quality Metric} (MQM) \cite{mqm} and (in-line with the present task) \textit{Direct Assessments} (DA) \cite{graham-etal-2013-da}. The hyper-parameters used for these models are exactly as described in \citet{comet}, excluding the following alterations:  we use XLM-RoBERTa large instead of base and we increase the feed-forward hidden sizes (from 2304 in the first layer and 1152 in the second to 3072 and 1536 hidden units, respectively). We also keep the embedding layer frozen and apply a layer-wise learning rate decay (as proposed in \citet{howard2018universal}) by which each transformer layer has a learning rate scaled at 0.95 times the rate of the layer above. By doing this, we hope that our metric generalizes better to new language pairs introduced this year.

\subsection{Translation Ranking Model}
While for the Estimators using a larger pretrained encoder seems to improve performance we found that for the Translation Ranking Model, larger pretrained encoders lead to training instability and an overall worse performance. For that reason we decided to keep the model proposed in \cite{comet} without any alteration. 
\section{Corpora}

Below we provide an outline of the various datasets used to train our models:

\subsection{HTER Corpora}
Our HTER corpus is a concatenation of two publicly available corpora: the QT21 corpus
and the APE-QUEST corpus.
While the QT21 corpus contains segments from the information technology and life sciences domains \cite{specia-etal_MTSummit:2017}, the APE-QUEST contains segments from the legal domain \cite{ive-etal-2020-post}. Concatenation of these two corpora gives a total of 211K tuples with source sentence, corresponding human-generated reference, MT hypothesis, and post-edited MT (PE). With regard to the language pairs in each corpus, QT21 covers: English to German (en-de), Latvian (en-lt) and Czech (en-cs), and
German to English (de-en); while APE-QUEST covers: English-Dutch (en-nl), English-French (en-fr), English-Portuguese (en-pt). Finally, the HTER score is obtained by calculating the translation edit rate (TER) \cite{Snover06astudy} between the MT hypothesis and the corresponding PE. By doing this, we were able to create a large HTER corpus covering several language pairs and different domains.

\subsection{MQM Corpus}
Our MQM corpus is an extension of the proprietary corpus presented in \citet{comet}. This internal data consists of customer support chat messages translated using a domain adapted MT model and their corresponding references (consisting of post-edited translations from earlier iterations of the MT systems). The data was then MQM-annotated according to the guidelines set out in \citet{burchardt}. Our final corpus contains 27K tuples from English into 15 different languages and/or dialects: German (en--de), Spanish (en--es), Latin-American Spanish (en--es-latam), French (en--fr), Italian (en--it), Japanese (en--ja), Dutch (en--nl), Portuguese (en--pt), Brazilian Portuguese (en--pt-br), Russian (en--ru), Swedish (en--sv), Turkish (en--tr), Polish (en--pl),  simplified Chinese (en--zh-CN), and Taiwanese Chinese (en--zh-TW).

\subsection{DA Corpora}
Every year, since 2008, the WMT News Translation shared task organizers collect human judgements in the form of DAs. Since 2017, due to a lack of annotators, these scores are mapped to relative rankings ({\small{DA}}RR). We take advantage of this data in two ways: 1) we use the scores directly in order to train an estimator model, 2) as in \citet{comet}, we use the {\small{DA}}RR to train a translation ranking system. The collective corpora of 2017, 2018 and 2019 contain a total of 24 language pairs, including low-resource languages such as English-Gujarati (en-gu) and English-Kazakh (en-kk). For the purposes of this paper we use the data from 2017 and 2018 to train and the data from 2019 to validate. Later, for participation in the 2020 shared task, we intend to include the data from 2019 in our training corpus.

\section{Segment-Level Task}
\label{sec:seg-lvl-task}

\begin{table*}[!ht]
\centering
\caption{Segment-level Kendall's Tau ($\tau$) correlations for language pairs from English-to-other for the WMT19 Metrics {\footnotesize DA}RR corpus.}
\label{tab:english-to-x2019-seg}
\begin{tabular}{lccccccccc}
\hline
 & \textbf{en-cs} & \textbf{en-de} & \textbf{en-fi} & \textbf{en-gu} & \textbf{en-kk} & \textbf{en-lt} & \textbf{en-ru} & \multicolumn{1}{c|}{\textbf{en-zh}} & \\
 Nº Tuples  & 27178 & 99840 & 31820 & 11355 & 18172 & 17401 & 24334 & \multicolumn{1}{c|}{18658} & \textbf{avg.} \\\specialrule{1.5pt}{1pt}{1pt}
{\sc Bleu}& 0.364 & 0.248 & 0.395 & 0.463 & 0.363 & 0.333 & 0.4691 & \multicolumn{1}{c|}{0.235} & 0.410  \\
{\sc chrF} & 0.444 & 0.321 & 0.518 & 0.548 & 0.510 & 0.438 & 0.548 & \multicolumn{1}{c|}{0.241} & 0.510  \\
{\sc Bertscore} (F1) & 0.486 & 0.350 & 0.526 & 0.559 & 0.534 & 0.464 & 0.581 & \multicolumn{1}{c|}{0.350} & 0.550 \\
{\sc Prism} & 0.580 & 0.416 & 0.590 & - & 0.529 & 0.555 & 0.581 & \multicolumn{1}{c|}{0.373} & 0.518  \\ \hline
{\sc Comet-mqm} (large) & 0.595 & 0.405  & 0.594 & 0.580 & 0.546 & 0.607  & 0.693 & \multicolumn{1}{c|}{0.400} & 0.553 \\
{\sc Comet-hter} (large) & 0.610 & 0.427 & 0.610 & 0.587 & 0.569 & 0.615 & 0.707 & \multicolumn{1}{c|}{0.405} & 0.566 \\
{\sc Comet-da} (large)  & \textbf{0.618} & \textbf{0.435} & 0.620 & \textbf{0.617} & 0.585 & 0.619 & \textbf{0.711} & \multicolumn{1}{c|}{0.427} & 0.579 \\
{\sc Comet-rank} (base)  & 0.603 & 0.427 & \textbf{0.664} & 0.611 & \textbf{0.693} & \textbf{0.665} & 0.580 & \multicolumn{1}{c|}{\textbf{0.449}} & \textbf{0.587} \\
\end{tabular}
\end{table*}

\begin{table*}[!ht]
\centering
\caption{Segment-level Kendall's Tau ($\tau$) correlations on language pairs with English as a target for the WMT19 Metrics {\footnotesize DA}RR corpus.}
\label{tab:x-to-english2019-seg}
\begin{tabular}{lcccccccc}
\hline
  & \textbf{de-en} & \textbf{fi-en} & \textbf{gu-en} & \textbf{kk-en} & \textbf{lt-en} & \textbf{ru-en} & \multicolumn{1}{c|}{\textbf{zh-en}} &  \\
Nº Tuples  & 85365 & 32179 & 20110 & 9728 & 21862 & 39852 & \multicolumn{1}{c|}{31070} & \textbf{avg.} \\ \specialrule{1.5pt}{1pt}{1pt}
{\sc Bleu}& 0.054 & 0.236 & 0.194 & 0.276 & 0.249 & 0.115 & \multicolumn{1}{c|}{0.321} & 0.206  \\
{\sc chrF} & 0.123 & 0.292 & 0.240 & 0.323 & 0.304 & 0.177 & \multicolumn{1}{c|}{0.371} & 0.261  \\
{\sc Bertscore} (F1) & 0.191 & 0.354 & 0.292 & 0.351 & 0.381 & 0.221 &\multicolumn{1}{c|}{0.433} & 0.318 \\
{\sc Bleurt} (large-512) & 0.174 & 0.374 & 0.313 & 0.372 & 0.388 & 0.220 &\multicolumn{1}{c|}{ 0.436} & 0.325 \\
{\sc Prism}  & 0.189 & 0.366 & 0.320 & 0.362 & 0.382 & 0.220 & \multicolumn{1}{c|}{0.434} & 0.325     \\ \hline
{\sc Comet-mqm} (large) & 0.191 & 0.360 & 0.289 & 0.346 & 0.373 & 0.213 & \multicolumn{1}{c|}{0.426} & 0.314    \\
{\sc Comet-hter} (large) & 0.193 & 0.351 & 0.286 & 0.340 & 0.375 & 0.209 & \multicolumn{1}{c|}{0.429} & 0.312     \\
{\sc Comet-da} (large) & \textbf{0.220} & 0.368 & 0.316 & \textbf{0.378} & \textbf{0.405} & \textbf{0.231} & \multicolumn{1}{c|}{\textbf{0.462}} & \textbf{0.340}  \\
{\sc Comet-rank} (base)  & 0.202 & \textbf{0.399} & \textbf{0.341} & 0.358 & \textbf{0.407} & 0.180 & \multicolumn{1}{c|}{0.445} & 0.333
\end{tabular}
\end{table*}

At segment-level, we take each of our Estimator models trained to predict MQM, HTER and DA and predict segment-level scores on the {\small{DA}}RR data from WMT19. We then generate pairwise rankings based on these predicted scores. For each language pair we apply the formulation of Kendall's Tau ($\tau$) from the shared task \cite{metrics-2019-results} as follows:

\begin{equation}
    \label{eq:kendall}
    \tau = \frac{\textit{Concordant} - \textit{Discordant}}{\textit{Concordant} + \textit{Discordant}}
\end{equation}

\textit{Concordant} here being the number of times a metric assigns a higher score to the ``better'' hypothesis $h^+$ and \textit{Discordant}, the number of times a metric assigns a higher score to the ``worse'' hypothesis $h^-$, or that the evaluation was otherwise equal.

\section{Document-Level Task}
\label{sec:doc-lvl-task}
In the WMT2019 News Translation the organizers introduced a document-level translation task \cite{barrault-etal-2019-findings} for en-de and en-cs. This means that for those language pairs we are able to obtain document-level direct assessments. We can compute a score taking into account an entire document and correlate it with the human evaluation also carried out at document-level. 

For our document-level submission we propose the generation of a document-level score as a weighted average of the predicted scores for each segment composing that document (hereinafter called micro-average score), where the same is weighted by segment length. 

To calculate this score at inference time we pass the entire document (divided into segments) through the model as a single batch. This has the added effect of reducing inference time.

\section{System-level Task}

Following previous years, the metric used in the System-level Task will be Pearson's $r$ correlation score. The correlation is calculated between the average of all DA human z-scores for a given system and language pair, and the average of the corresponding scores predicted by a given metric. Because the goal of some metrics is to maximize the correlation with human judgements (i.e. {\sc Bleu}), while for others is to minimize that correlation (i.e. HTER), the value reported is its absolute value. 

\subsection{Robustness to high-permorming systems}
One important finding from WMT19 is the general deterioration of metrics' performance when considering only the top $n$ MT systems \cite{metrics-2019-results}. Previously, we showed robustness of our metrics in this scenario in terms of Kendall's Tau at segment-level \citep{comet}. \citet{mathur2020tangled} show that at system-level, Pearson correlation is highly influenced by outliers and that performances for metrics such as {\sc Bleu} drop significantly when considering only the top systems. To address this, we propose a system-level pairwise comparison measured with the same Kendall's Tau formulation used for segment-level analysis outlined in section \ref{sec:seg-lvl-task} above. By doing this, we are not only better handling possible outliers, but emulating a real world application of these metrics: In most cases (both in academia and industry), we want a metric that can successfully differentiate between two systems, even if those systems are very close in terms of performance, which is often the case.

\section{Quality Estimation as a Metric}
Given the clear parallels between the {\sc Comet} framework and modern approaches to Quality Estimation such as \citet{kepler2019unbabels}, we used our framework to participate in the ``QE as a Metric'' track of the shared task by removing the reference at input and proportionately reducing the dimensions of the feed-forward network to accommodate the reduced input.

\section{Multi-Reference Handling}
\label{sec:multi-ref}
In this year's shared task we are provided with a second human-generated reference for German-to-English (de-en), Russian-English (ru-en) and Chinese-to-English (zh-en). Given that our base framework currently supports the input of only one single reference, we introduce a method of leveraging information from a second reference at inference time.

During standard training of our models, we input source, hypothesis and reference in that order, resulting in a concatenation of embeddings as detailed further in \citet{comet}. During training, with a probability of $p=0.5$ we switch the positions of source and reference, such that the system receives the reference as the source and vice versa. This has two primary effects on our model. Firstly, during fine-tuning of the underlying language model, the source embeddings are aligned with the target language embedding space resulting in more useful source embeddings. Secondly, it forces the model to treat source and reference as interchangeable inputs, allowing it to handle switching of inputs at inference time without excessively hindering the model's predictive ability. Finally, at inference time we embed source $\bm{s}$, hypothesis $\bm{h}$, reference $\bm{r}$ and the alternative reference $\bm{\hat{r}}$. These embeddings are then passed to the feed-forward neural network in the following six permutations: $[\bm{s};\bm{h};\bm{r}]$, $[\bm{r};\bm{h};\bm{s}]$, $[\bm{s};\bm{h};\bm{\hat{r}}]$, $[\bm{\hat{r}};\bm{h};\bm{s}]$, $[\bm{r};\bm{h};\bm{\hat{r}}]$ and $[\bm{\hat{r}};\bm{h};\bm{r}]$. 

Six passes through the feed-forward gives us six predictions. Final, aggregated scores are achieved by taking the mean of the six predictions and multiplying it by 1 minus the standard deviation ($\sigma$). The intuition being that $1-\sigma$ gives something of an idea of confidence of the model at the segment-level and that scaling the mean prediction to penalize lower confidence might align better with human judgement.

\begin{table*}[!ht]
\centering
\caption{Kendall's Tau ($\tau$) correlation and standard deviation ($\sigma$) across all language pairs for the top 5 high-performing systems.}
\label{tab:kendall-sys}
\begin{tabular}{lcc}
Model              & \multicolumn{1}{l}{Avg. Kendall (all)} & \multicolumn{1}{l}{Avg. Kendall (en)} \\ \specialrule{1.5pt}{1pt}{1pt}
{\sc Bleu}              & 0.387$\pm$0.366 & 0.257$\pm$0.395   \\
{\sc chrF}               & 0.387$\pm$0.463 & 0.343$\pm$0.513   \\
{\sc Bertscore} (F1)     & 0.453$\pm$0.267 & 0.429$\pm$0.279   \\
{\sc Bleurt}            & -               & 0.571$\pm$0.355   \\ 
{\sc Prism}              & 0.52$\pm$0.270  & 0.514$\pm$0.279   \\ \hline
{\sc Comet-mqm} (large)   & 0.587$\pm$0.277 & 0.543$\pm$0.276   \\
{\sc Comet-hter} (large)  & 0.547$\pm$0.325 & 0.486$\pm$0.363   \\
{\sc Comet-da} (large)    & \textbf{0.653}$\pm$\textbf{0.233} & \textbf{0.629}$\pm$\textbf{0.269}  \\
{\sc Comet-rank} (base) & 0.547$\pm$0.256 & 0.543$\pm$0.276  \\

\end{tabular}
\end{table*}

\section{Experimental Results}
\label{sec:length}
Below we present results of our various {\sc Comet} models on WMT19 evaluation sets as described above. Segment-level and document-level results are outlined in the corresponding tables within the body of the paper, the remaining tables for other task results are contained in the Appendices hereto.

\subsection{Segment-level Task}

Our segment-level results on the shared task test sets for WMT19 are detailed in tables \ref{tab:english-to-x2019-seg} and \ref{tab:x-to-english2019-seg}. We note that for all language pairs out of English (Table \ref{tab:english-to-x2019-seg}) both our DA Estimator and our {\sc Comet-rank} (base) outperform prior metrics, in some cases by a significant margins. The same can be said in most language pairs into English, where we consistently perform at the level competitive with or exceeding prior metric performance in this task. Table \ref{tab:not-english2019-seg} (contained in the appendices) further illustrates performance of our models on non-English language pairs. We note that in all settings our {\sc Comet} models outperform state-of-the-art for these language pairs.

\subsection{System-level Task}

System-level results are outlined in tables \ref{tab:english-to-x2019-sys}, \ref{tab:x-to-english2019-sys} and \ref{tab:not-english2019-sys} in the appendix. In most language pairs we outperform the best metrics in terms of correlation with human judgement. For those language pairs for which our metrics are outperformed by others, we note that ours are at least competitive with other, recent metrics. 

An unexpected result is that at system-level our {\sc Comet-rank} (base) does not perform as well as our Estimators, regardless of its strong segment-level results. We believe that this is an artifact of training directly on {\small{DA}}RR data. Since in WMT shared tasks, the DA rating scale employed is defined at the 0-25-50-75-100 point margins, the minimum required difference between two hypothesis to produce {\small{DA}}RR judgement is 25 points \cite{metrics-2019-results}. All other segments are discarded, as within that range the notion of which hypothesis is better becomes ambiguous. As a result we believe that our ranker model learns to successfully discriminate 
less ambiguous examples and struggles to correctly assign a score otherwise.

\subsubsection{Robustness to high-performing systems}

As outlined above, we also complement our evaluation at system-level with an analysis of metric performance in terms of the pairwise ranking of the top five performing systems from each language pair. For each setting we output the Kendall's Tau (that is to say the formulation outlined in section \ref{sec:seg-lvl-task} above) and report the mean and standard deviation of results across language pairs in table \ref{tab:kendall-sys}.

In both settings we note that our DA Estimator (large) model significantly outperforms other metrics both in terms of mean and standard deviation. This strongly suggests that not only do we perform well in terms of system-level Pearson but that at a practical level, our model can much more successfully differentiate high-performing systems.

\subsection{Document-level Task}
Table \ref{fig:micro-vs-macro} compares the micro-averaging against a simple unweighted average. From table \ref{fig:micro-vs-macro} we can observe that micro-averaging outperforms macro-averaging by a small margin. Table \ref{tab:doc-level} summarizes our results for the Document-level Task using our segment-level Estimators with micro-averaging. In this task, the HTER Estimator shows generally superior performance on average surpassing our best performing segment-level model, the DA Estimator. An important conclusion to draw from the strong document-level correlations noted here is that a model trained to generate segment-level scores, can also perform well as a document-level metric.

\begin{table}[!ht]
\centering
\caption{Pearson correlation ($r$) between Document-level DAs and micro average segment-level scores for English-to-German and English-to-Czech.}
\label{tab:doc-level}
\begin{tabular}{lccc}
\hline
  & \textbf{en-cs} & \textbf{en-de} & \\ 
Nº Documents & 1115 & \multicolumn{1}{c|}{2355} &  \textbf{avg.} \\ \specialrule{1.5pt}{1pt}{1pt} 
{\sc Comet-mqm} (large) & 0.638 &  \multicolumn{1}{c|}{0.516} & 0.577 \\
{\sc Comet-hter} (large) & 0.655 &  \multicolumn{1}{c|}{\textbf{0.558}} & \textbf{0.607} \\
{\sc Comet-da} (large) & \textbf{0.667} &  \multicolumn{1}{c|}{0.528} & 0.598 \\
\end{tabular}
\end{table}

\begin{table}[!ht]
\centering
\caption{Pearson correlations ($r$) and adequacy (as reported in \citet{freitag-bleu-paraphrase-references-2020}) for segment-level DA using our DA Estimator (large) model on WMT19 Metrics shared task test data for en-de. We show Pearson's $r$ for the single reference scenario using the corresponding reference (`1-ref') and the multi-reference scenario where the reference is combined with the original in the manner outlined in section \ref{sec:multi-ref} above (`2-ref').}
\label{tab:multiref}
\begin{tabular}{llll}
\hline
\textbf{Reference} & \textbf{Adequacy} & \textbf{$r$ (1-ref)} &  \textbf{$r$ (2-ref)} \\ 
\specialrule{1.5pt}{1pt}{1pt} 
WMT & 85.3 & 0.523 & - \\
AR & 86.7 & 0.539 & 0.555 \\
WMTp & 81.8 & 0.470 & 0.529 \\
ARp & 80.8 & 0.476 & 0.537 \\
\end{tabular}
\end{table}
\subsection{Multi-Reference Handling}
\label{sec:multi-ref-res}
Additional references were obtained for two language pairs: en-de and de-en. For the former, we conducted experiments using 3 additional references from \citet{freitag-bleu-paraphrase-references-2020}: AR reference (an additional high quality reference translation), ARp reference (a paraphrased-as-much-as-possible version of AR), and WMTp reference (a paraphrased-as-much-as-possible version of the original WMT reference); for the latter, we use the alternative reference given in the WMT19 News shared task test set. Conveniently, \citet{freitag-bleu-paraphrase-references-2020} also offer a notion of the quality of the extra references for en-de by providing human-generated adequacy assessments for each. In table \ref{tab:multiref} we show the performance of our DA Estimator (large) model with each reference, either as a single reference or combined in the manner described in section \ref{sec:multi-ref} above with the original reference.

\newcommand{\hwplotA}{\raisebox{2pt}{\tikz{\draw[blue,solid,line width=1.2pt](0,0) -- (5mm,0);}}}
\newcommand{\hwplotB}{\raisebox{2pt}{\tikz{\draw[red,solid,line width=1.2pt](0,0) -- (5mm,0);}}}
\newcommand{\hwplotC}{\raisebox{2pt}{\tikz{\draw[brown,solid,line width=1.2pt](0,0) -- (5mm,0);}}}
\begin{figure}[ht!]
\centering
\resizebox{\columnwidth}{!}{%
\begin{tikzpicture}
\begin{axis}[
    ybar,
    enlargelimits=0.35,
    title= {Segment-level Kendall's Tau ($\tau$)},
    symbolic x coords={AR ref.,ARp ref.,WMTp ref.},
    y tick label style={/pgf/number format/precision=5},
    xtick=data,
    ticklabel style = {font=\small},
    ]
\addplot coordinates {
    (AR ref., 0.4483173077) 
    (ARp ref.,0.4416266026) 
    (WMTp ref.,0.4344751603)
};
\addplot coordinates {
    (AR ref.,0.4507612179) 
    (ARp ref.,0.4426282051) 
    (WMTp ref.,0.4363982372)
};
\addplot coordinates {
    (AR ref.,0.4420472756) 
    (ARp ref.,0.4420472756) 
    (WMTp ref.,0.4420472756)
};
\end{axis}
\end{tikzpicture}
}
\caption{Performance impact of using different kinds of references in combination with the original WMT English-to-German reference. In {\hwplotA} we observe the Kendall-Tau $\tau$ ranking correlation achieved by our multi-reference Estimator model (section \ref{sec:multi-ref}). In {\hwplotB} we present the Kendall-Tau $\tau$ ranking correlation of our ``one-reference'' Estimator model using the alternative reference. Finally, for comparison, in {\hwplotC} we show the Kendall-Tau $\tau$ ranking correlation of our  ``one-reference'' Estimator model using the original reference.} 
\label{fig:multi-ref-kendall-en-de} 
\end{figure}
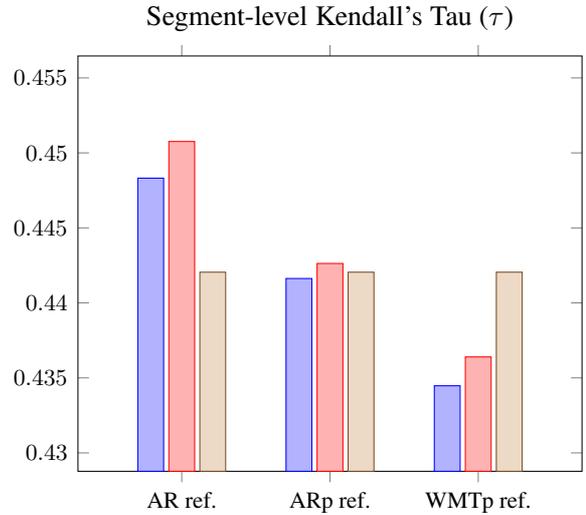

While we lack data to draw any statistically significant conclusions, there is a strong suggestion from these results of a positive correlation between reference quality and utility to the predictive model.

For de-en, using an alternative reference did not offer any gain in Pearson's $r$. We note that when using it alone we only achieve $r$=0.34 compared to using the original reference which achieves $r$=0.42. We speculate, based on our observations above, that this might be due to the alternative reference being of lower quality. 

These results potentially show that for approaches such as {\sc Comet}, quality is more important than quantity, and that lower-quality additional references can potentially hurt rather than help improve the correlations obtained using only one single high-quality reference.

With regard to the Kendall Tau measured at segment-level, by looking at Figure \ref{fig:multi-ref-kendall-en-de} (en-de), we see no significant differences in using the multi-reference technique. This suggests that having a higher Pearson's $r$ score does not necessarily guarantee a better Kendall's Tau. 

We note that by design, with an approach such as {\sc Comet} that is based on a meaning-representation of references, extra references are expected to provide only minor additional value, especially versus lexical-based metrics such as {\sc Bleu} \cite{papineni-etal-2002-bleu}.  Whereas the adequacy of the reference(s) is (again by design) expected to have a more significant impact on the performance of the model.  Our initial results seem to strongly support this hypothesis.

\section{Conclusions}

In this paper we present {\sc Comet}, Unbabel's contribution to the WMT 2020 Metrics shared task. We leverage the framework outlined in \citet{comet} to demonstrate state-of-the-art or otherwise competitive levels of correlation with human judgements in all tasks and introduce a novel method of making optimal use of alternative references and demonstrate that the quality of the reference used is relevant to the success of our framework. Further investigation of the latter, in particular how to better leverage different kinds of references, represent an interesting direction for future work.


\section{Acknowledgments}
We are grateful to Fabio Kepler, Daan Van Stigt, Miguel Vera, and the reviewers, for their valuable feedback and discussions. This work was supported in part by the P2020 Program through projects MAIA and Unbabel4EU, supervised by ANI under contract numbers 045909 and 042671, respectively.

\bibliography{anthology,emnlp2020}
\bibliographystyle{acl_natbib}

\appendix

\section{Appendix A}

\label{sec:appendix}

\begin{table*}[!ht]
\centering
\caption{Segment-level Kendall's Tau ($\tau$) correlations on language pairs not involving English for the WMT19 Metrics {\small DA}RR corpus. {\sc Comet-rank} (base) scores are to be replaced with results of the large model.}
\label{tab:not-english2019-seg}
\begin{tabular}{lcccc}
\hline
  & \textbf{de-cs} & \textbf{de-fr} & \multicolumn{1}{c|}{\textbf{fr-de}} & \\ 
Nº Tuples & 23194 & 4862 & \multicolumn{1}{c|}{1369} & \textbf{avg.} \\ \specialrule{1.5pt}{1pt}{1pt} 
{\sc Bleu}& 0.222 & 0.226 &  \multicolumn{1}{c|}{0.173} & 0.207 \\
{\sc chrF} & 0.341 & 0.287 &  \multicolumn{1}{c|}{0.274} & 0.301 \\
{\sc Bertscore} (F1) & 0.356 & 0.330 &  \multicolumn{1}{c|}{0.277} & 0.321 \\
{\sc Prism}   & 0.452 & 0.443 &  \multicolumn{1}{c|}{\textbf{0.421}} & 0.439  \\ \hline
{\sc Comet-mqm} (large) & 0.413 & 0.422 &  \multicolumn{1}{c|}{0.327} & 0.387 \\
{\sc Comet-hter} (large) & 0.425 & 0.449 &  \multicolumn{1}{c|}{0.381} & 0.418 \\
{\sc Comet-da} (large) & \textbf{0.471} & \textbf{0.469} & \multicolumn{1}{c|}{\textbf{0.420}} & \textbf{0.453} \\
{\sc Comet-rank} (base)  & 0.389 & 0.444 & \multicolumn{1}{c|}{0.331} & 0.388 \\
\end{tabular}
\end{table*}

\begin{table*}[!ht]
\centering
\caption{System-level Pearson correlation ($r$) for the from-English language pairs from WMT19 DA corpus. {\footnotesize{DA}}RR Ranker (base) scores are to be replaced with results of the large model.}
\label{tab:english-to-x2019-sys}
\begin{tabular}{lccccccccc}
\hline
  & \textbf{en-cs} & \textbf{en-de} & \textbf{en-fi} & \textbf{en-gu} & \textbf{en-kk} & \textbf{en-lt} & \textbf{en-ru} & \multicolumn{1}{c|}{\textbf{en-zh}} \\
  Nº Systems  & 11 & 22 & 12 & 11 & 10 & 12 & 12 & \multicolumn{1}{c|}{12} &  \textbf{avg.} \\
  \specialrule{1.5pt}{1pt}{1pt}
{\sc Bleu}& \textbf{0.988} & 0.952 & 0.978 & 0.780 & 0.864 & 0.979 & 0.973 & \multicolumn{1}{c|}{0.762} & 0.910  \\
{\sc chrF} & \textbf{0.986} & 0.983 & \textbf{0.988} & 0.839 & 0.969 & 0.964 & 0.979 & \multicolumn{1}{c|}{0.822} & 0.941  \\
{\sc Bertscore} (F1) & 0.983 & 0.990 & 0.969 & 0.907 & 0.983 & 0.972 & \textbf{0.989} & \multicolumn{1}{c|}{0.927} & 0.965  \\
{\sc Prism} & 0.964 & 0.987 & 0.947 & -     & 0.978 & 0.929 & 0.914 & \multicolumn{1}{c|}{0.900} & 0.946  \\ \hline
{\sc Comet-mqm} (large) & 0.943 & 0.968 & 0.949 & 0.946 & 0.979 & \textbf{0.985} & 0.966 & \multicolumn{1}{c|}{0.958} & 0.962 \\
{\sc Comet-hter} (large) & 0.948 & \textbf{0.991} & 0.959 & 0.948 & 0.965 & \textbf{0.982} & 0.973 & \multicolumn{1}{c|}{0.943} & 0.964 \\
{\sc Comet-da} (large)  & 0.964 & \textbf{0.995} & 0.969 & \textbf{0.964} & \textbf{0.989} & \textbf{0.982} & \textbf{0.987} & \multicolumn{1}{c|}{\textbf{0.969}} & \textbf{0.977}  \\
{\sc Comet-rank} (base)  & 0.943 & 0.937 & 0.914 & 0.817 & 0.963 & 0.973 & 0.861 & \multicolumn{1}{c|}{0.942} & 0.919 \\
\end{tabular}
\end{table*}

\begin{table*}[!ht]
\centering
\caption{System-level Pearson correlation ($r$) for the into-English language pairs from WMT19 DA corpus. {\footnotesize{DA}}RR Ranker (base) scores are to be replaced with results of the large model.}
\label{tab:x-to-english2019-sys}
\begin{tabular}{lcccccccc}
\hline
   & \textbf{de-en} & \textbf{fi-en} & \textbf{gu-en} & \textbf{kk-en} & \textbf{lt-en} & \textbf{ru-en} & \multicolumn{1}{c|}{\textbf{zh-en}}  \\ 
Nº Systems  & 16 & 11 & 9 & 7 & 11 & 13 & \multicolumn{1}{c|}{15} &  \textbf{avg.} \\
\specialrule{1.5pt}{1pt}{1pt}
{\sc Bleu}& 0.879 & 0.984 & 0.975 & 0.959 & 0.969 & 0.840 & \multicolumn{1}{c|}{0.895} & 0.929 \\
{\sc chrF} & 0.916 & \textbf{0.988} & 0.967 & 0.982 & 0.938 & 0.942 & \multicolumn{1}{c|}{0.952} & 0.955 \\
{\sc Bertscore} (F1) & 0.949 & 0.984 & \textbf{0.990} & \textbf{0.995} & 0.961 & 0.901 & \multicolumn{1}{c|}{0.982} & 0.966 \\
{\sc Bleurt} (large-512) & 0.939 & 0.984 & 0.989 & 0.989 & \textbf{0.992} & \textbf{0.980} & \multicolumn{1}{c|}{\textbf{0.994}} & \textbf{0.981} \\
{\sc Prism} & \textbf{0.954} & 0.981 & \textbf{0.992} & \textbf{0.992} & \textbf{0.994} & 0.905 & \multicolumn{1}{c|}{\textbf{0.992}} & 0.973 \\ \hline
{\sc Comet-mqm} (large)  & 0.926 & 0.974 & 0.972 & 0.971 & 0.986 & 0.889 & \multicolumn{1}{c|}{0.959} & 0.954    \\
{\sc Comet-hter} (large) & 0.918 & 0.953 & 0.958 & 0.951 & 0.983 & 0.924 & \multicolumn{1}{c|}{0.978} & 0.952     \\
{\sc Comet-da} (large)   & 0.946 & 0.983 & \textbf{0.993} & \textbf{0.996} & \textbf{0.993} & 0.970 & \multicolumn{1}{c|}{\textbf{0.993}} & \textbf{0.982}  \\
{\sc Comet-rank} (base)  & 0.922 & 0.981 & 0.963 & 0.932 & 0.987 & 0.674 & \multicolumn{1}{c|}{0.967} & 0.918 \\
\end{tabular}
\end{table*}

\begin{table*}[!ht]
\centering
\caption{System-level Pearson correlation ($r$) for language pairs not involving English from WMT19 DA corpus.}
\label{tab:not-english2019-sys}
\begin{tabular}{lcccc}
\hline
  & \textbf{de-cs} & \textbf{de-fr} & \multicolumn{1}{c|}{\textbf{fr-de}} & \\ 
Nº Systems & 9 & 11 & \multicolumn{1}{c|}{10} & \textbf{avg.} \\ \specialrule{1.5pt}{1pt}{1pt}
{\sc Bleu}  & 0.936 & 0.934 & \multicolumn{1}{c|}{0.869} & 0.913  \\
{\sc chrF}   & \textbf{0.994} & 0.933 & \multicolumn{1}{c|}{0.908} & 0.945  \\
{\sc Bertscore} (F1)   & 0.988 & 0.953 & \multicolumn{1}{c|}{0.942} & 0.961  \\ 
{\sc Prism}   & 0.988 & 0.924 & \multicolumn{1}{c|}{0.922} & 0.945  \\ \hline
{\sc Comet-mqm} (large) & 0.936 & 0.950 & \multicolumn{1}{c|}{0.885} & 0.924 \\
{\sc Comet-hter} (large) & 0.951 & 0.901 & \multicolumn{1}{c|}{0.924} & 0.925 \\
{\sc Comet-da} (large) & 0.973 & \textbf{0.972} & \multicolumn{1}{c|}{\textbf{0.954}} & \textbf{0.966} \\
{\sc Comet-rank} (base)  & 0.819 & 0.941 & \multicolumn{1}{c|}{0.927} & 0.896 \\
\end{tabular}
\end{table*}

\begin{table*}[!ht]
\centering
\caption{Document-level Pearson correlation ($r$) for micro average and macro average for English-to-German and English-to-Czech.}
\label{fig:micro-vs-macro}
\begin{tabular}{lcccc}\hline
           & \multicolumn{2}{c}{\textbf{en-cs}} & \multicolumn{2}{c}{\textbf{en-de}} \\
           & Micro-avg.  & Macro-avg.  & Micro-avg.  & Macro-avg.  \\ \specialrule{1.5pt}{1pt}{1pt} 
{\sc Comet-da} (large) & \textbf{0.667}      & \multicolumn{1}{c|}{0.660}      & 0.528      & \textbf{0.529}      \\
{\sc Comet-mqm} (large) & 0.638      & \multicolumn{1}{c|}{\textbf{0.639}}      & 0.516      & \textbf{0.519}      \\
{\sc Comet-hter} (large)  & \textbf{0.655}       & \multicolumn{1}{c|}{0.650}      & \textbf{0.558}      & 0.552       \\\hline
  & \textbf{0.653}      & \multicolumn{1}{c|}{0.649}      & \textbf{0.534}       & 0.533     
\end{tabular}
\end{table*}


\end{document}